\documentclass[letterpaper, 10 pt, conference]{ieeeconf}  

\IEEEoverridecommandlockouts                              

\usepackage{graphicx}
\usepackage{multicol}
\usepackage{multirow}
\usepackage{array}
\usepackage{booktabs}
\usepackage{amsmath} 
\usepackage{amssymb}  
\usepackage{adjustbox} 
\title{\LARGE \bf
Adaptive Neural Networks for Intelligent Data-Driven Development
}

\author{Youssef Shoeb$^{\dagger,\ddagger}$, Azarm Nowzad$^{\dagger}$ and Hanno Gottschalk$^{\ddagger}$
\thanks{*The research leading to these results is funded by the German Federal Ministry for Economic Affairs and Climate Action within the project “just better DATA".
}
\thanks{$^{\dagger}$ Continental AG, Germany,
        }%
\thanks{$^{\ddagger}$ Technische Universität Berlin, Germany
        }%
}

\begin{document}

\maketitle
\thispagestyle{empty}
\pagestyle{empty}

\begin{abstract}
Advances in machine learning methods for computer vision tasks have led to their consideration for safety-critical applications like autonomous driving.
However, effectively integrating these methods into the automotive development lifecycle remains challenging.
Since the performance of machine learning algorithms relies heavily on the training data provided, the data and model development lifecycle play a key role in successfully integrating these components into the product development lifecycle.
Existing models frequently encounter difficulties recognizing or adapting to novel instances not present in the original training dataset. This poses a significant risk for reliable deployment in dynamic environments. To address this challenge, we propose an adaptive neural network architecture and an iterative development framework that enables users to efficiently incorporate previously unknown objects into the current perception system.
Our approach builds on continuous learning, emphasizing the necessity of dynamic updates to reflect real-world deployment conditions.
Specifically, we introduce a pipeline with three key components: (1) a scalable network extension strategy to integrate new classes while preserving existing performance, (2) a dynamic OoD detection component that requires no additional retraining for newly added classes, and (3) a retrieval-based data augmentation process tailored for safety-critical deployments.
The integration of these components establishes a pragmatic and adaptive pipeline for the continuous evolution of perception systems in the context of autonomous driving.
\end{abstract}

\section{INTRODUCTION}
Computer vision systems have greatly improved due to recent advancements in machine learning (ML) algorithms. Where traditional approaches relied on hand-crafted rules and features, ML algorithms learn complex mapping functions directly from training data. This data-driven paradigm, powered by advances in neural architectures and computational resources, has outperformed traditional computer vision methods in various applications such as autonomous driving, medical imaging, and industrial inspection~\cite{plaksyvyi2023comparative}. While this shift from explicit rule-based approaches to data-driven systems has enabled remarkable progress, it also introduces significant challenges, particularly in ensuring the functional safety of automated driving systems~\cite{9256519,9984982}.

Some of the main challenges in developing ML-based systems include identifying critical variables that affect model performance and establishing robust performance guarantees for various scenarios. To address these challenges, researchers have developed different methodologies for safely building systems with ML components\cite{waymo_v_model,iterative_v_model}.
In these approaches, strong emphasis is placed on ensuring data coverage for a well-defined operational domain design (ODD), followed by carefully orchestrated model development and testing processes.
However, even with well-curated datasets, real-world deployment can expose models to unanticipated, out-of-distribution (OoD) inputs that degrade the performance of machine learning algorithms, a critical concern in safety-sensitive domains like autonomous driving~\cite{shoeb2025out}.

\begin{figure}[t]
      \centering
      \includegraphics[width=1\linewidth]{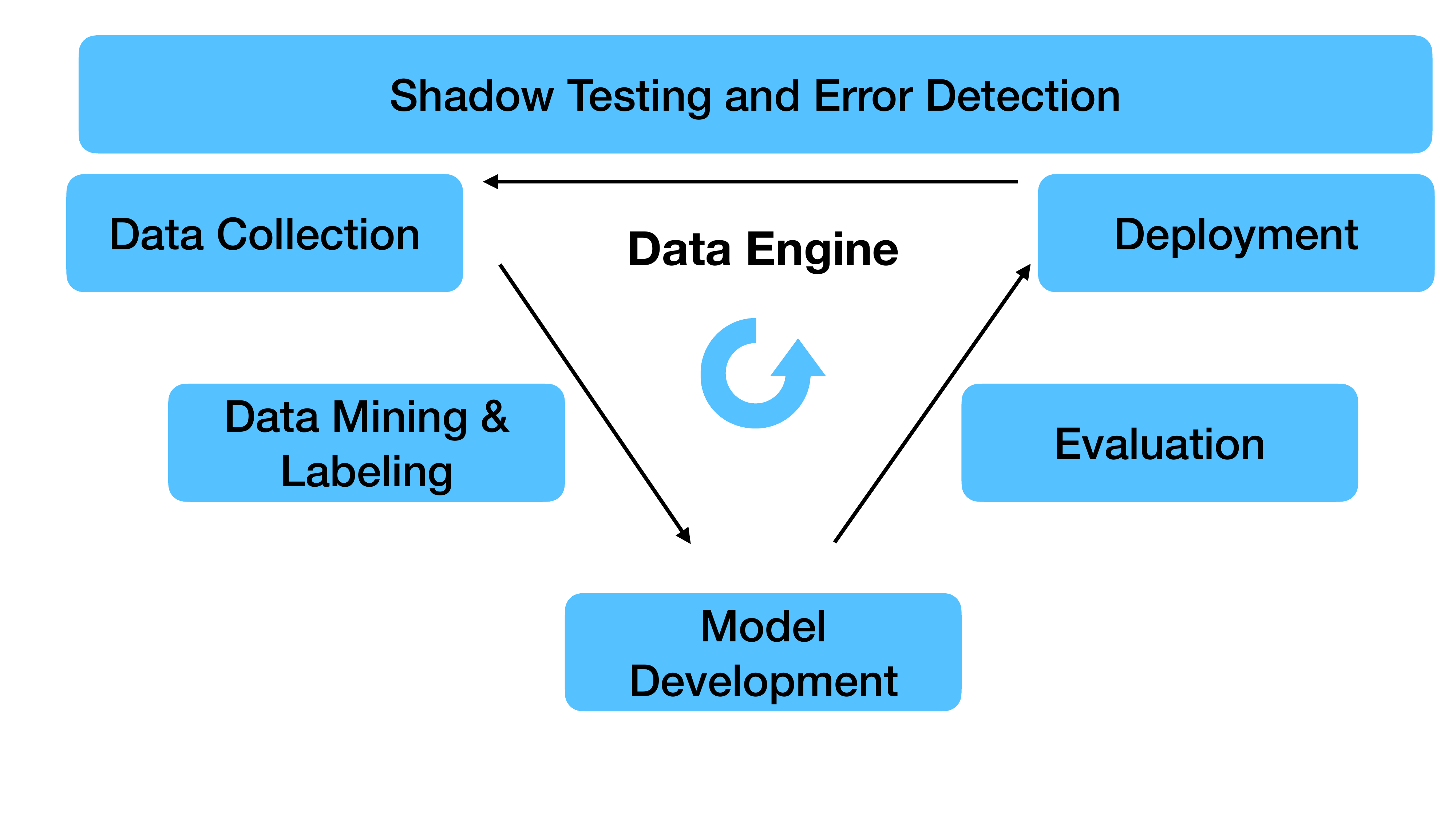}
      \caption{V-Model structure of the Data Engine, illustrating the key stages of data collection, model development, deployment, and shadow testing/error detection within a fully data-driven machine learning methodology.}
      \label{data_engine}
   \end{figure}

To systematically address OoD challenges, prior work has introduced the data engine concept~\cite{iterative_v_model} (see Fig.~\ref{data_engine}), an iterative approach that continuously refines datasets and updates models based on observed distributions.
However, this approach relies on batch-based model retraining, requiring rigorous evaluation and testing to ensure functional safety compliance.
This limitation underscores the necessity for a more adaptable and efficient method that dynamically permits perception models to incorporate new object categories without requiring comprehensive model retraining.
Building on this idea, we propose a novel pipeline for detecting and learning OoD objects in autonomous driving scenarios.
Our approach iteratively adapts the underlying model with minimal computational overhead, ensuring efficient real-time updates based on user-defined requirements.
Understanding distribution shifts is critical for ensuring robustness in perception models. We classify these shifts into two main categories: (1) domain shift, where familiar objects appear under different conditions (e.g., altered sensor settings), and (2) semantic shift, where entirely new object categories emerge in the environment (e.g., electric scooters).
While challenging, domain shifts are often anticipated and partially addressed through comprehensive operational domain design and data collection processes, which aim to capture various environmental variations.
On the other hand, semantic shifts represent a fundamental mismatch between the model's learned categories and the objects encountered in real-world scenarios.
Therefore, semantic shifts cannot be fully anticipated during the training phase and often require a mechanism to alert the model when new object categories arise and to learn them to incorporate these new classes into the environment model. In this work, we attempt to address some of the challenges to resolve some of the issues with semantic shift. 

We present a modular extension strategy that enables the incremental integration of new object classes while preserving base model performance. Unlike standard transfer learning, which requires fine-tuning and may degrade previously learned representations, our approach facilitates selective adaptation without catastrophic forgetting. Specifically, we introduce a parameter-efficient extension mechanism that allows a perception model to dynamically incorporate new categories based on user needs, enabling a customizable and user-centric perception system.
We propose a retrieval-based data augmentation process that enhances adaptation to novel objects. Instead of relying solely on manually labelled datasets for retraining, our framework leverages a structured retrieval system to select relevant samples from previously encountered instances. This ensures that new class integration does not suffer from domain shift and remains data-efficient and computationally scalable. 
Our approach enables incremental expansion of object detection capabilities according to a user's specific application requirements.

\section{SMART DATA LOOP FOR LEARNING NEW CLASSES}

   \begin{figure}[thpb]
      \centering
      \includegraphics[width=\linewidth]{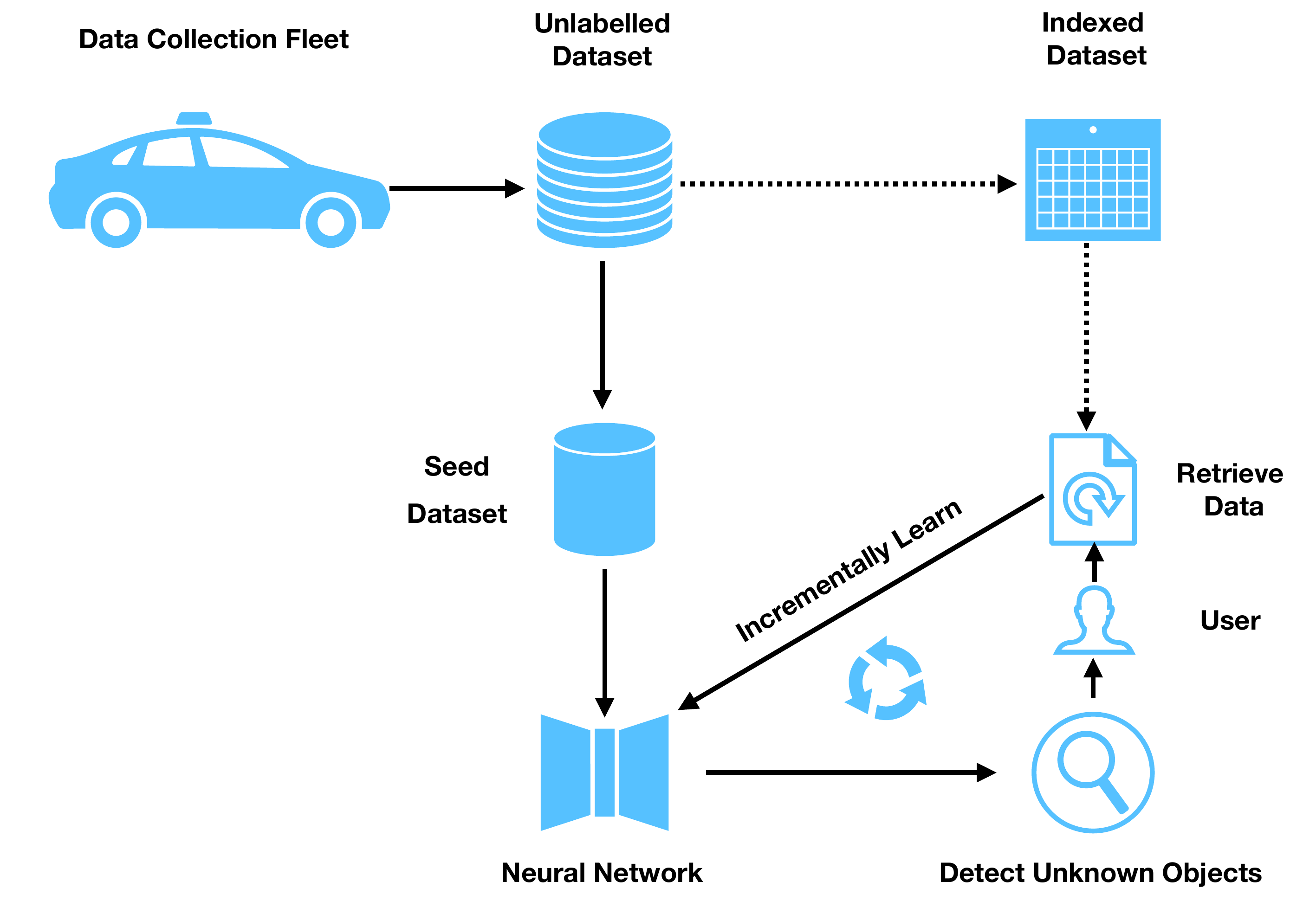}
      \caption{Development cycle for a dynamic perception system.}
      \label{overview}
   \end{figure}

\subsection{Overview}

Figure~\ref{overview} provides a high-level overview of the proposed development pipeline for an ML-based perception system capable of learning OoD objects incrementally based on a user's input. In the first stage of ML product development, a data collection campaign starts with a fleet of vehicles recording data. Human experts then label a large amount of data to create a labelled subset of the data, commonly referred to as the seed dataset.
The seed dataset is split into training, validation, and test sets, which are used to build and evaluate the model.
The model is trained on the training set; the validation set is used to avoid overfitting and help design the network architecture, and the test set's performance is used to verify the network and determine the expected performance during deployment.
For such a construction to be valid, the data distributions must be identical; this is typically satisfied by collecting enough data and randomly splitting the training, validation and test sets.
However, encountering distribution shifts and other errors is inevitable during deployment in open-world applications like autonomous driving. 

To allow for user-specific customization needs, the perception systems must be able to detect and adapt to objects not seen in the training data, while maintaining performance on already learned classes.
The standard safety approach for OoD instances is to treat all OoD objects as unknown and maintain a higher safety margin, as their behaviour is unknown.
However, during deployment, it might be necessary to learn new classes as they become recurring elements in regular traffic scenarios or for personalized adaptation for specialized environments.
Existing methods for handling such scenarios typically require manual relabeling and full model retraining, making them inefficient for dynamic settings.
To address this, we propose an adaptive class expansion framework that dynamically incorporates newly encountered objects by coupling a dynamic GMM-based classification network with an OoD detection module and a retrieval-based decision mechanism.

Our proposed framework consists of three key components: a generative segmentation architecture that measures class-conditional densities, an OoD detection module, a retrieval-based augmentation method, and a generative-based learning approach. 
First, a generative architecture is used for dense prediction of the objects in the image. Then, an OoD detection module identifies previously unseen objects and flags them as potential candidates for expanding the model. Once OoD objects are detected, the system allows the user to specify which objects to learn, after which the system retrieves similar instances from a large-scale unlabelled dataset to facilitate learning. When a new class is selected for integration, an adaptive classification head is dynamically added to the network; leveraging the generative-based classifier, the model is able to contrastively learn new categories while preserving previously learned ones, mitigating the risk of catastrophic forgetting. 
By integrating these three components, our framework automates the discovery of new concepts while keeping a human in the loop for confirmation. This ultimately forms a smart data curation loop that improves model adaptability with minimal manual intervention.

\subsection{Initial Segmentation Network}
   \begin{figure}[thpb]
      \centering
      \includegraphics[width=\linewidth]{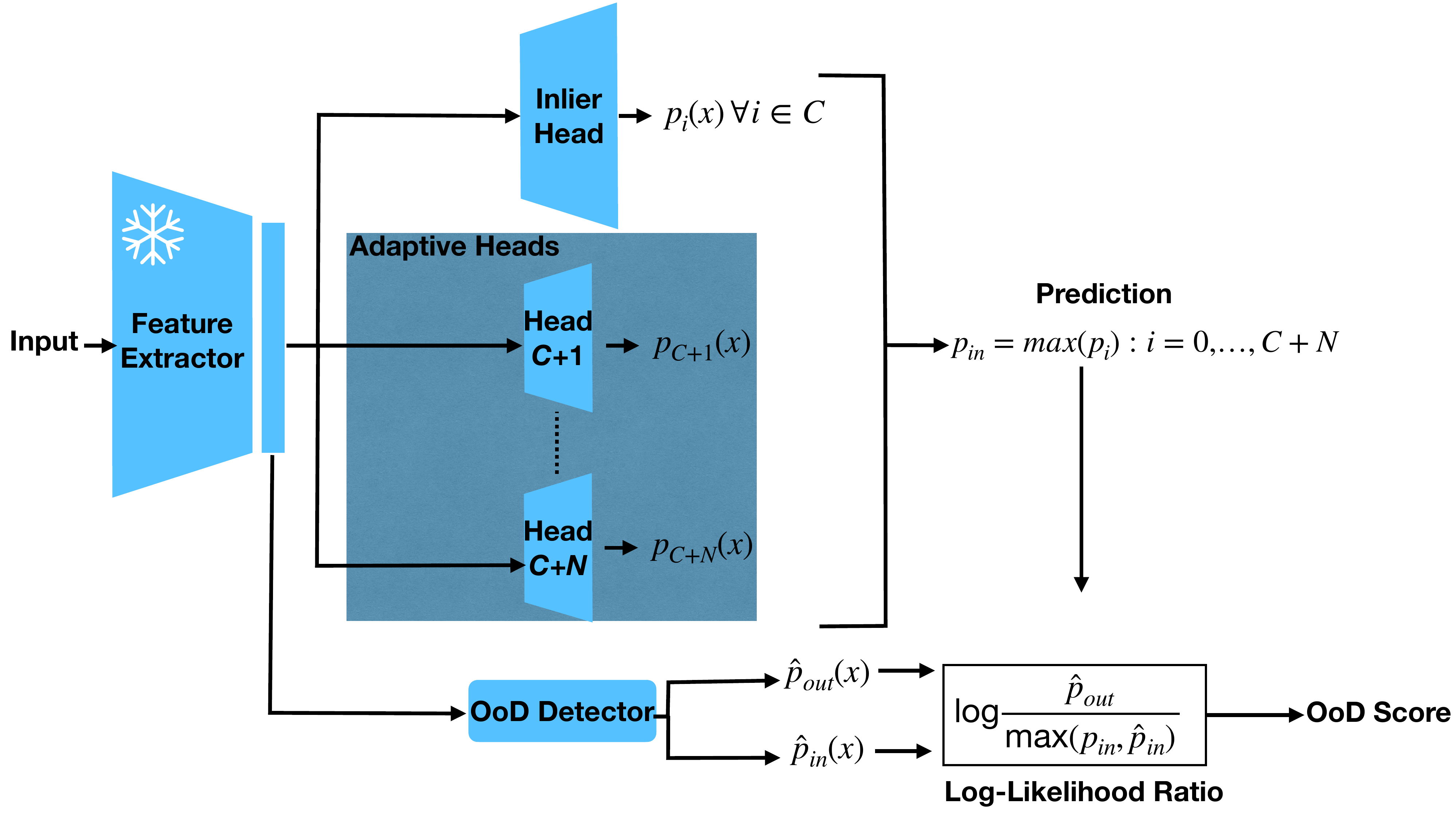}
      \caption{Adaptive neural network architecture with an OoD detection mechanism.}
      \label{segmentation_network}
   \end{figure}

For pixel-level segmentation, we utilize the DinoV2~\cite{Oquab2024TMLR} feature encoder, which is a self-supervised ViT \cite{dosovitskiy2021an} that has been shown to produce robust and rich visual representations for multiple tasks~\cite{Ranzinger2024CVPR}. To maintain this rich representation, the feature encoder is frozen during all stages of training. DinoV2 downscales the spatial resolution by a factor of fourteen. To upscale the features to the high resolution, we utilise a Feature Pyramid Network (FPN)~\cite{Lin2017CVPR} that takes features from multiple encoder layers and fuses them to produce a high-resolution output feature map. The classification head of the segmentation network is a set of Gaussian Mixture Models (GMMs) where each class is represented as a separate GMM with a uniform prior on the component weights:
\begin{equation}
    p(x | y, \theta) = \sum_{c=1}^{C}{\pi_{yc}~\mathcal{N}(x ; \mu_{yc}, \Sigma_{yc})}
\end{equation}
where $C$ is the number of components per GMM, $\pi_{yc}$ is the component mixture weight for component $c$ of class $y$, $\mu_{yc}$,$\Sigma_{yc}$ are the mean and covariance matrix respectively, and $\mathcal{N}$ is the Gaussian distribution.
The parameters of the decoder $\theta$ are optimized by minimizing the cross-entropy loss on the training set $\mathcal{D}_{in}$ over the posterior output of the parameters of the GMM $\phi$:
\begin{equation}
      \theta^*=\text{argmin}_{\theta}-\sum\nolimits_{(x,y)\in\mathcal{D}_{in}}\log p(y|x;\{\phi_y\}_{y=1}^{|Y|}, \theta),
\end{equation}
In contrast to conventional softmax classifiers, which make hard categorical decisions, this approach explicitly models the class-conditional probability distributions. By leveraging GMMs for classification, we obtain a probabilistic measure of class membership, which allows for more flexible class integration when new categories emerge.
The GMM parameters are optimized using a Sinkhorn-based Expectation-Maximization (EM) algorithm, which differs from standard EM by enforcing balanced assignment constraints. This ensures that each mixture component learns a well-distributed feature representation, preventing mode collapse and improving generalization when adding new classes. 
We refer the reader to \cite{Liang2022NeurIPS} for more details on the training strategy.

\subsection{OoD Detection}
Current state-of-the-art OoD detection methodologies depend on some sort of outlier supervision to learn cues for detecting OoD objects. Typically, in a fine-tuning step, the parameters of an inlier model are adjusted to give either a high-entropy or low-logit score to a set of known unknown objects. OoD objects are then detected by thresholding the entropy or the logit scores of the model~\cite{Nayal2023ICCV}. 
However, this method encounters challenges in incremental learning settings, where the in-distribution space dynamically expands. With each newly learned class, the fine-tuning step must be repeated.

To address this issue, we propose a likelihood ratio-based formulation that adapts to evolving in-distribution classes without requiring explicit re-training of the OoD detection module.
Inspired by Nayal et al.~\cite{nayal2024likelihoodratiobasedapproachsegmenting}, our method ensures that the OoD score appropriately adjusts to newly incorporated classes, preventing them from being detected as OoD once they have been learned.

We include annotated OoD objects as known unknowns and train an OoD detection module that consists of a 3-layer Multi-Layer Perceptron (MLP) and a generative classifier to learn two classes: an OoD distribution $\hat{p}_{out}(x)$ and a generic inlier distribution $\hat{p}_{in}(x)$ learned directly from the features of the backbone. The OoD score for a pixel is then:
\begin{equation}
    \text{log}\left( \frac{\hat{p}_{out}(x)}{\text{max}\left (\hat{p}_{in}(x), p_{in}(x)\right)} \right)
\end{equation}
where $p_{in}(x)$ is the density of the most likely inlier class:
\begin{equation}
    p_{in}(x) = \text{max}_k\, \text{log} \left ( p(k|x)\right )
\end{equation}
where $k$ includes both original classes from the inlier decoder and any additional heads that were incrementally added (check Fig. \ref{segmentation_network}). 

Compared to~\cite{nayal2024likelihoodratiobasedapproachsegmenting}, the advantage of our approach is that it adaptively accounts for newly included in-distribution classes. Specifically, if $p_{in}(x)$ becomes sufficiently large for one of the newly learned classes, instances belonging to this class will no longer be flagged as OoD. This ensures that our method remains robust in dynamic scenarios where the in-distribution space expands over time—without requiring retraining of the OoD detection module.

\subsection{Retrieval}
To enable effective search within a large-scale unlabeled dataset, we propose a retrieval framework that allows users to find relevant samples based on either semantic descriptions or visual references. Different retrieval needs depend on the user's use case. When a user is searching for instances based on general object categories—such as “trailer” or “e-scooter”—a text-based retrieval approach is ideal. In such cases, joint vision-language embeddings enable semantic retrieval, allowing users to discover relevant objects even when exact matches are unavailable~\cite{shoeb2024have}.
Conversely, when a user has a specific reference object and wants to retrieve visually similar instances—such as different viewpoints of the same anomaly—an image-based retrieval method typically provides a more precise search mechanism based purely on visual similarity.

The text-based retrieval approach relies on CLIP~\cite{radford2021learning}, which encodes images and textual inputs into a shared high-dimensional embedding space. CLIP’s image encoder first processes all detected OoD objects in the unlabeled dataset to produce feature vectors, which are then indexed using an approximate nearest neighbour (ANN) structure. This ANN index enables sublinear retrieval by pruning the search space through heuristics or quantization. When a user enters a text query, we encode it with CLIP’s text encoder to generate a query embedding and retrieve the top-$n$ objects that maximize the cosine similarity $\text{sim}(z,v_i)$. Formally, this can be written as:
\begin{equation}
    \text{top-}n \approx \arg\max_{1 \le i \le N} \text{sim}\bigl(z,\,v_i\bigr),
\end{equation}
where $\{v_i\}_i^{N}$ are the indexed feature vectors and $z$ is the query embedding. 
Because the embeddings are precomputed and the search is performed in the ANN index, this method significantly reduces computational overhead compared to brute-force retrieval, thereby enabling real-time results on large-scale datasets.

Our second approach follows a similar structure but uses embeddings generated by the DINOv2-based backbone used in the segmentation network, allowing queries to be made with images rather than text. Before retrieval, we cluster the DINOv2 embeddings of the detected OoD objects to group visually similar objects, which helps users select representative images that depict the target concept or object.
These chosen objects are then encoded into the same embedding space, and we once again utilize the ANN index to retrieve the top-$n$ closest matches based on cosine similarity. By relying on visual rather than textual queries, this approach offers an alternative that does not require the additional text model for retrieval.

To ensure scalability with increasing dataset size, we employ an online $k$-Means approach, where centroids are updated progressively as new data points arrive. The centroid update follows an exponential moving average (EMA) update rule:
\begin{equation}
    C_{j}^{(t+1)} = (1-\alpha)C_{j}^{(t)} + \alpha x
\end{equation}
where $C_{j}^{(t)}$ is the centroid of cluster $j$ at iteration $t$, $x$ is the newly assigned data point, and $\alpha$ is the learning rate.
Before updating the centroid, each new data point is assigned to its nearest cluster. 
This method avoids recomputing clusters from scratch, making it suitable for dynamic datasets.

In practice, after obtaining the results of a query, we apply a post-processing step to eliminate low-similarity matches, which may occur when the dataset does not contain $n$ valid instances of the queried object. The threshold is set based on the retrieval method: for image-based, we observe that the cosine similarity between correct samples falls between 0.7 and 1, while for text-based retrieval, correct matches are found in the lower range of 0.25 and 0.38, due to the modality gap in CLIP~\cite{ModalityGap} which affects the alignment between vision and language representations. This filtering step removes samples that fail to meet a similarity threshold, ensuring that the retrieved set remains representative of the user’s initial query and is applicable for use in learning.

\subsection{Continual Learning}
In the practical application of ML deployments, updates to models may be constrained due to limitations in certification and safety guarantees. In such cases, standard continual learning methods such as weight regularization~\cite{michieli2019incremental} or data replay-based methods~\cite{recall_based} become infeasible as they inherently rely on modifying the network parameters. For our use case, we utilize a modular architecture that introduces new parameters to enable continual learning without compromising previously learned knowledge.  

The main challenge then becomes ensuring that newly learned classes are sufficiently separated from the features of already learned classes. Specifically, when new classes are introduced, their learned feature representations may overlap with those of previously learned classes, causing the model to misclassify previously learned classes during inference. Therefore, we use the already learned GMMs for each of the learned classes as prototypes to maintain class boundaries and prevent drift in feature space, ensuring that new classes are learned in distinct regions. To achieve this, we freeze all parameters of the inlier model and train only the extended adaptive heads. The adaptive heads are composed of a lightweight version of the FPN decoder used in the inlier segmentation but with 16 features instead of 256. The classification head is trained with a contrastive term that enforces separation from previously learned classes. The final loss function for each adaptive head becomes:
\begin{equation} 
\begin{split}
\mathcal{L} &= \underbrace{-\frac{1}{N} \sum_{i=1}^{N} \log p(y_i | x_i, \phi_{\text{new}})}_{\text{inlier loss}} \\&- \underbrace{\frac{\lambda }{N} \sum_{i=1}^{N} \log \frac{p(x_i | y_i, \phi_{\text{new}})}{\sum_{y' \in \mathcal{Y}} p(x_i | y', \phi_{\text{old}})}}_{\text{contrastive loss}}
\end{split}
\end{equation}
where $\lambda$ is a weighting coefficient, and  $p(y_i | x_i, G_{\text{new}})$ represents the posterior probability from the new head's classifier, and the contrastive denominator aggregates probabilities over all previously learned classes. This contrastive loss ensures that each feature in the image is assigned to a unique component within the GMM of the ground truth class.

\section{EVALUATIONS}

\subsection{Experimental Setup}
We demonstrate our workflow with the following scenario: starting with a model initially trained on the Cityscapes dataset~\cite{cordts2016cityscapes}, we run inference on the Street Obstacle Sequences (SOS) and Wuppertal Obstacle Sequences (WOS) datasets \cite{maag2022two}. 
These two datasets include 16 objects not included in Cityscapes's training data.
We first evaluate the model's performance in detecting these previously unknown objects.
We then evaluate how well the two proposed retrieval methods work in retrieving the detected OoD objects given either an image or text query.
Finally, we evaluate how well the incrementally learned model performs after training on the retrieved data.  

\subsubsection{Metrics}
To assess the inlier segmentation performance, we rely on the standard Mean Intersection over Union (mIoU) metric $=\frac{1}{N}\sum_{i}^{N} \frac{\text{TP}_i}{\text{TP}_i+\text{FP}_i+\text{FN}_i}$, where TP$_i$, FP$_i$, and FN$_i$ are the numbers of true positive, false positive, and false negative pixels for class i, respectively, and $N$ is the number of classes. For the OoD detection, we consider the component-level F1 Score, where a component is defined as a set of connected pixels.
A ground-truth segment is considered a true positive if the component-wise IoU is greater than 0.25 with any predicted components. False positives are identified as predicted OoD components with an IoU less than 0.25 with any ground-truth OoD segments.
We evaluate the retrieval performance by measuring how often the top-ranked prediction exactly matches the ground-truth label when used as a query.
Specifically, for each ground-truth class $c$, we calculate the proportion of images of class $c$ whose predicted label with the highest similarity score is also $c$. This proportion is often referred to as top-1 accuracy for class $c$. We then take the mean of these values across all classes, yielding a top-1 macro-precision $mp = \frac{1}{|C|} \sum_{c \in C} \frac{TP_c}{TP_c+FP_c}$. This allows us to assess how well the model returns relevant results higher than irrelevant ones when queried with the ground truth labels.

\subsection{Results}
\begin{table}[h]
\centering
\caption{Comparison of OoD detection methods across SOS and WOS, and the inlier performance on Cityscapes}
\begin{tabular}{|l|c|c|c|c|}
\hline
\textbf{Method} & \textbf{Dataset} & \textbf{F1 Score} & \textbf{mIoU} & \textbf{Num. Parameters} \\
\hline
\multirow{2}{*}{ours} & SOS & 80.23 & \multirow{2}{*}{80.3} & \multirow{2}{*}{31.6M} \\
\cline{2-3}
 & WOS & 54.44 & & \\  
\hline
\multirow{2}{*}{UEM~\cite{nayal2024likelihoodratiobasedapproachsegmenting}} & SOS & 85.62 & \multirow{2}{*}{81.1} & \multirow{2}{*}{102.0M} \\
\cline{2-3}
 & WOS & 56.55 & & \\ 
\hline
\multirow{2}{*}{RbA~\cite{Nayal2023ICCV}} & SOS & 77.28 & \multirow{2}{*}{82.8} & \multirow{2}{*}{203.7M} \\ 
\cline{2-3}
 & WOS & 58.26 & & \\
\hline
\end{tabular}

\label{tab:method_comparison}
\end{table}
\subsubsection{Inlier and OoD Detection Performance}
Table~\ref{tab:method_comparison} compares our approach with two recent state-of-the-art methods (UEM~\cite{nayal2024likelihoodratiobasedapproachsegmenting} and RbA~\cite{Nayal2023ICCV}) across both datasets. Averaged across both datasets, our method performs 0.05 less than RbA and 3.8 less than UEM in the F1 scores. However, our method requires significantly fewer parameters: 3.2 times fewer than UEM and 6.4 times fewer than RbA. This reduction in model size makes our approach more suitable for resource-constrained environments, such as in-car systems, where available memory and computational capacity are limited. Moreover, the efficiency of our model allows for faster inference, making it a viable choice for real-time applications where low latency is critical.

\subsubsection{Retrieval Performance}

\begin{table}[h]
\centering
\caption{Retrieval Performance on SOS and WOS}
\begin{tabular}{|l|c|c|c|}
\hline
\textbf{Method} & \textbf{Dataset} & \textbf{mp} & \textbf{Num. Parameters} \\
\hline
\multirow{2}{*}{Image-Retrieval} & SOS & 79.76  & \multirow{2}{*}{22.1M} \\
\cline{2-3}
 & WOS & 63.08 & \\ 
\hline
\multirow{2}{*}{Text-Retrieval} & SOS & 68.28 & \multirow{2}{*}{149.6M} \\ 
\cline{2-3}
 & WOS & 65.06 & \\
\hline
\end{tabular}

\label{tab:retrieval}
\end{table}

Table~\ref{tab:retrieval} shows the retrieval performance of the image-based and text-based retrieval on the SOS and WOS datasets. Averaged across the two datasets, the image-based retrieval performs better than the text-based retrieval by 4.75 mp, while having $\approx 7$ times less parameters. However, the image-based retrieval requires a clustering step to obtain the centroids, which the user can select as queries, in this examples we used simple $k$-means clustering on the embeddings to obtain the clusters, and the parameter $k$ was obtained using the elbow method. However, if the parameter $k$ is set to a value which is lower than the number of objects in the dataset, then some of the potential object queries might be missed. If the number of objects cannot be roughly estimated before clustering, then text-based retrieval might be a better option.

We also evaluate using only CLIP's image encoder for image-based retrieval. Compared to an equally sized DinoV2 model and averaged across both datasets, the performance of the DinoV2 results in $\approx 8$ mAP point improvement. This performance difference highlights the importance of choosing the appropriate embedding model for image-based retrieval tasks. Additionally, this suggests that self-supervised models like DinoV2 may better cluster visual features in certain settings.

\subsubsection{Continual Learning}
Table~\ref{tab:continual} shows the performance on each incrementally learned class and the original 19 classes. All retrieved instances are split into 80:20 splits for training and testing. The mIoU on the validation set of the newly learned class is less than the mIoU for the original inlier class with a margin of 2.5 mIoU points. This decrease in performance is most likely due to the nature of the newly learned objects being more difficult to learn due to the lack of sample points. Additionally, we find that our model performs better on larger objects for the original and incrementally added classes than on smaller objects. This discrepancy can be attributed to the network’s architecture, which only upsamples feature maps to 2/7 of the original image resolution, with the final output bilinearly upsampled by a factor of 3.5. This leads to the loss of fine details, making it more challenging for the model to classify smaller objects accurately.

A single adaptive head requires $\approx$ 1.3M parameters, which is $\approx 4\%$ of the original number of parameters. While this overhead remains manageable for a few additional classes, scalability becomes a concern as the number of learned classes increases. To address this, a batch update strategy, where multiple new classes are trained simultaneously rather than separately, can improve parameter efficiency by allowing shared features across classes. Additionally, regularization techniques or pruning strategies can be explored to reduce redundant parameters and mitigate memory constraints.

\begin{table*}[ht]
\centering
\caption{IoU results for the original 19 classes and the incrementally learned classes.}
\small
\begin{adjustbox}{max width=\textwidth}
\begin{tabular}{cccccccccc}
\toprule
\multicolumn{10}{c}{\textbf{IoU (Original 19 Classes)}} \\
\midrule
Road & Sidewalk & Building & Wall & Fence & Pole & Traffic Light & Traffic Sign & Vegetation & Terrain \\
98.2 & 85.7 & 93.2 & 58.3 & 66.3 & 66.6 & 71.2 & 80.7 & 92.9 & 65.8 \\
\midrule
Sky & Person & Rider & Car & Truck & Bus & Train & Motorcycle & Bicycle & mIoU \\
95.4 & 83.3 & 63.5 & 95.6 & 86.4 & 91.4 & 84.0 & 67.8 & 78.4 & 80.3 \\
\midrule
\multicolumn{10}{c}{\textbf{IoU (New Classes)}} \\
\midrule
Box & Chair & Crutch & Umbrella & Ball & Toy & Shopping Cart & Trash Can & Garden Gnome & Bag \\
89.7 & 82.9 & 54.6 & 85.2 & 79.8 & 73.8 & 84.4 & 95.3 & 67.6 & 86.8 \\
\midrule
Bottle & Bucket & Scooter & Dog & Skateboard & Concrete Barrier & mIoU &  &  &  \\
55.6 & 84.5 & 61.0 & 78.2 & 76.5 & 85.9 & 77.8 &  &  &  \\
\bottomrule
\end{tabular}
\end{adjustbox}

\label{tab:continual}
\end{table*}

\section{CONCLUSIONS AND FUTURE WORK}
In this paper, we propose an adaptive neural network architecture for autonomous vehicle perception systems that addresses some of the challenges of detecting and incorporating OoD objects.
We construct a practical framework for the data engine process, which detects, retrieves, and learns OoD objects.
Our framework enables perception models to dynamically adapt to new object categories while preserving existing performance. It uses a parameter-efficient mechanism that minimizes computational overhead. Additionally, it allows deployed neural networks to be customized for scenarios involving typically uncommon objects.
We experimentally evaluate separate components of our framework and compare them to other state-of-the-art methods, showing the overall efficacy of our proposed solution.

For future work, we aim to address the limitations of the current retrieval-based approach. While our current approach leverages a retrieval mechanism to incorporate new object instances from an existing dataset, a significant limitation is the potential scarcity of representative samples for emerging object categories. Therefore, in future work, we envision complementing our retrieval strategy with generative models capable of synthesizing photorealistic images. Using recent advances in generative models, we can automatically generate diverse training samples to fill gaps in our database.

\bibliographystyle{ieee_fullname}
{\small
\bibliography{bibliography}}

\end{document}